\documentclass[runningheads]{llncs}

\usepackage{graphicx}
\usepackage{comment}
\usepackage{amsmath,amssymb}
\usepackage{color}
\usepackage{caption}
\usepackage{multirow}
\usepackage{rotating}

\usepackage[pagebackref=true,breaklinks=true,colorlinks,bookmarks=false]{hyperref}

\newcommand{\X}{\mathbf{X}}

\renewcommand{\P}{\mathbf{P}}
\renewcommand{\L}{\mathbf{L}}

\newcommand{\n}{\mathbf{n}}

\newcommand{\nop}[1]{}

\newcommand\figref[1]{Fig.~\ref{#1}}
\newcommand\tabref[1]{Table~\ref{#1}}

\def\eg{\emph{e.g.}}
\def\ie{\emph{i.e.}}

\begin{document}

\pagestyle{headings}
\mainmatter
\def\ECCVSubNumber{890}

\title{Structured3D: A Large Photo-realistic Dataset for Structured 3D Modeling}

\titlerunning{Structured3D: A Large Photo-realistic Dataset for Structured 3D Modeling}
\author{
Jia Zheng\textsuperscript{1,2}$^{*\dag}$ \and
Junfei Zhang\textsuperscript{1}$^*$ \and
Jing Li\textsuperscript{2} \and
Rui Tang\textsuperscript{1} \and \\
Shenghua Gao\textsuperscript{2,3} \and
Zihan Zhou\textsuperscript{4}
}
\authorrunning{J. Zheng et al.}
\institute{
\textsuperscript{1}KooLab, Kujiale.com \quad \textsuperscript{2}ShanghaiTech University \\
\textsuperscript{3}Shanghai Engineering Research Center of Intelligent Vision and Imaging \\
\textsuperscript{4}The Pennsylvania State University \\
\url{https://structured3d-dataset.org}
}

\maketitle

\newcommand\blfootnote[1]{%
\begingroup
\renewcommand\thefootnote{}\footnote{#1}%
\addtocounter{footnote}{-1}%
\endgroup
}

\blfootnote{*: Equal contribution.}
\blfootnote{$\dag$: The work was partially done when Jia Zheng interned at KooLab, Kujiale.com.}

\begin{abstract}
Recently, there has been growing interest in developing \break learning-based methods to detect and utilize salient semi-global or global structures, such as junctions, lines, planes, cuboids, smooth surfaces, and all types of symmetries, for 3D scene modeling and understanding. However, the ground truth annotations are often obtained via human labor, which is particularly challenging and inefficient for such tasks due to the large number of 3D structure instances (\eg, line segments) and other factors such as viewpoints and occlusions. In this paper, we present a new synthetic dataset, Structured3D, with the aim of providing large-scale photo-realistic images with rich 3D structure annotations for a wide spectrum of structured 3D modeling tasks. We take advantage of the availability of professional interior designs and automatically extract 3D structures from them. We generate high-quality images with an industry-leading rendering engine. We use our synthetic dataset in combination with real images to train deep networks for room layout estimation and demonstrate improved performance on benchmark datasets.
\end{abstract}

\section{Introduction}

Inferring 3D information from 2D sensory data such as images and videos has long been a central research topic in computer vision. Conventional approach to building 3D models typically relies on detecting, matching, and triangulating local image features (\eg, patches, superpixels, edges, and SIFT features). Although significant progress has been made over the past decades, these methods still suffer from some fundamental problems. In particular, local feature detection is sensitive to a large number of factors such as scene appearance (\eg, textureless areas and repetitive patterns), lighting conditions, and occlusions. Further, the noisy, point cloud-based 3D model often fails to meet the increasing demand for high-level 3D understanding in real-world applications.

\begin{figure}[t]
\centering
\begin{tabular}{ccc}
\includegraphics[height=1.2in]{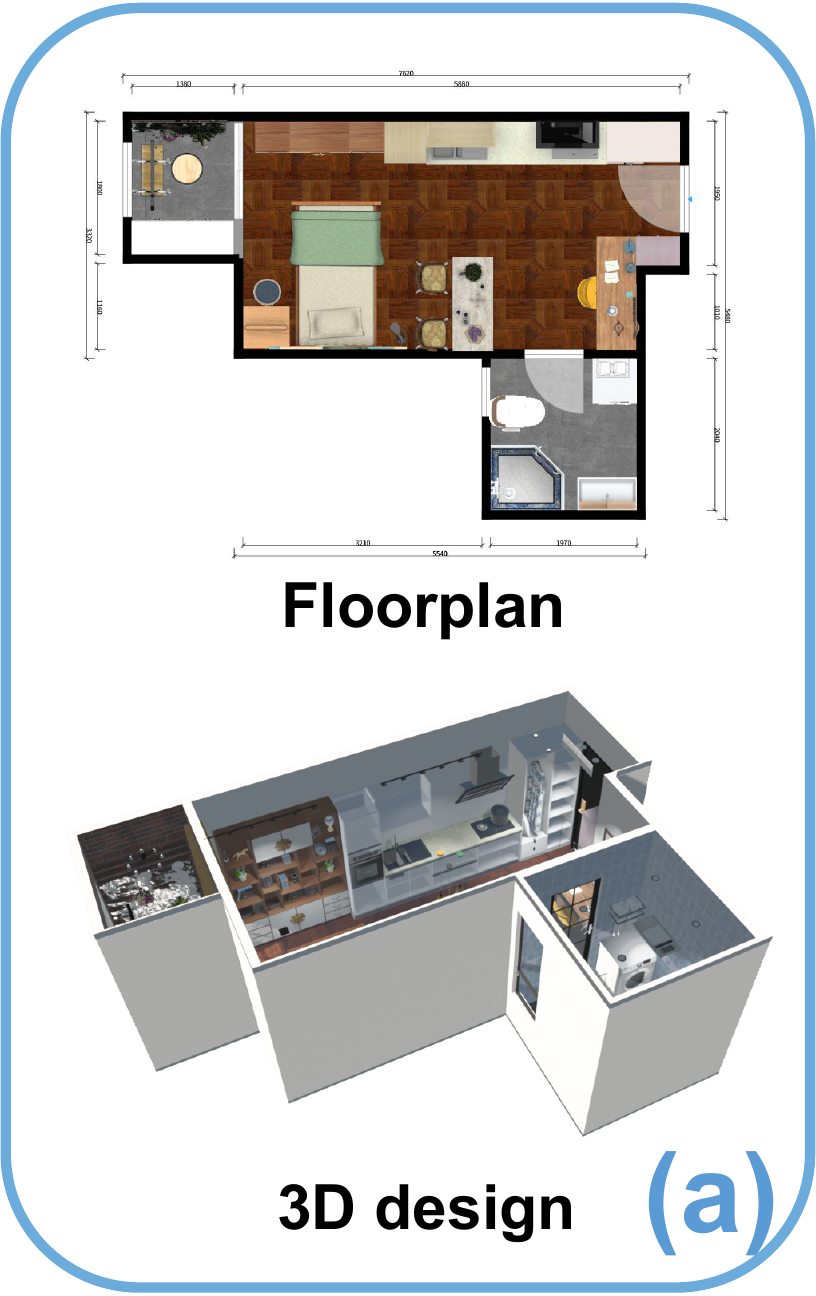}
\includegraphics[height=1.2in]{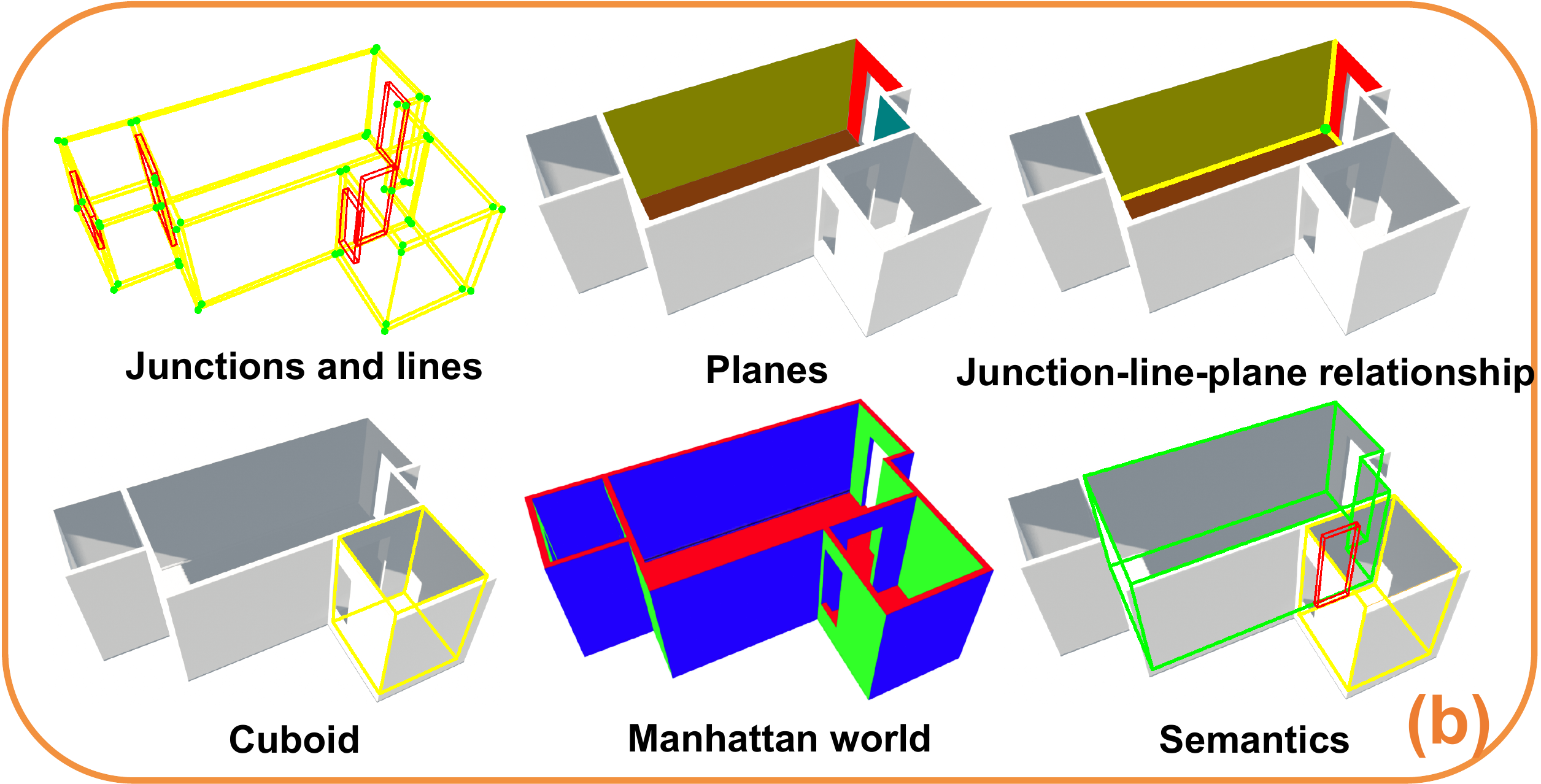}
\includegraphics[height=1.2in]{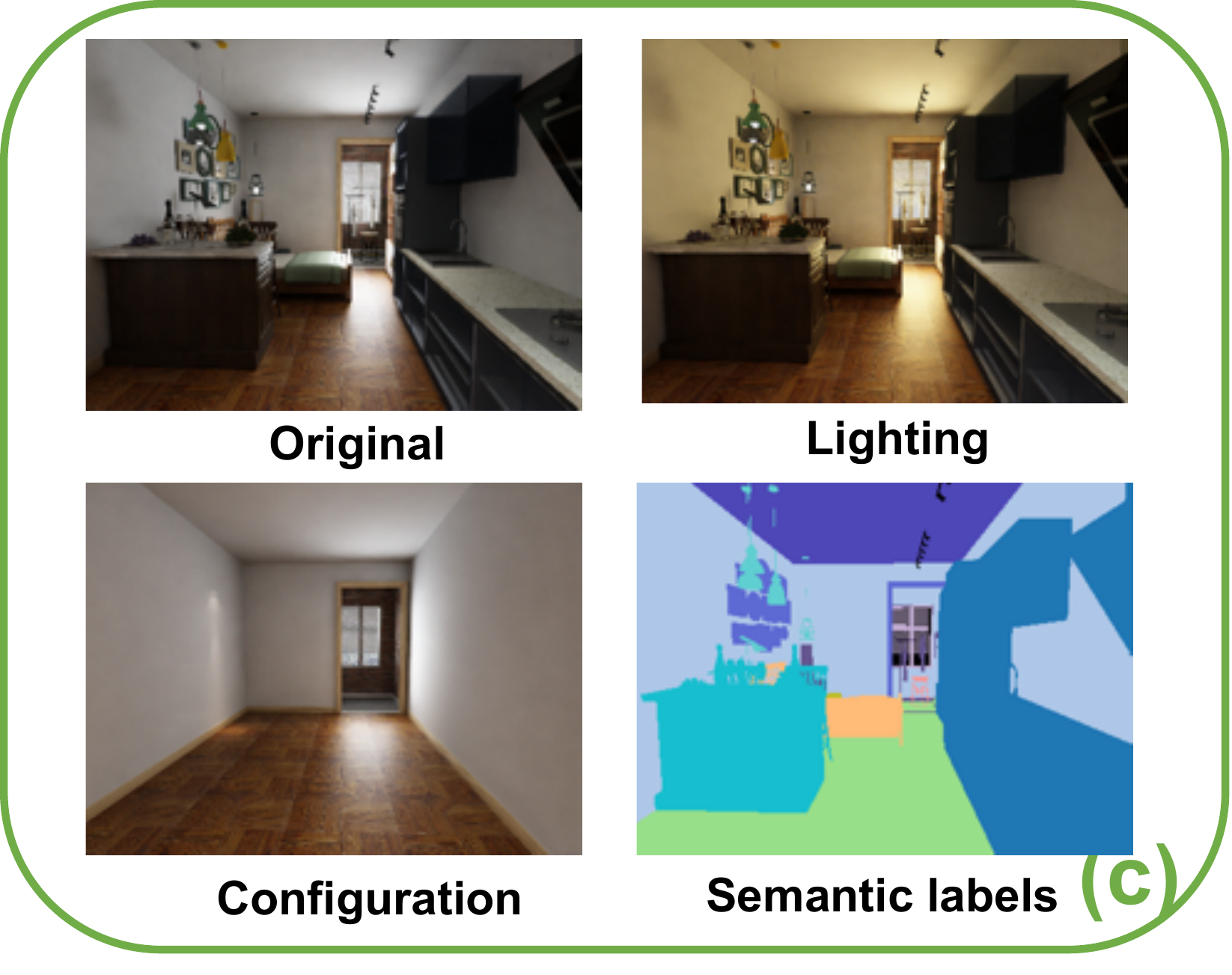}
\end{tabular}
\caption{The Structured3D dataset. From a large collection of house designs {\bf (a)} created by professional designers, we automatically extract a variety of ground truth 3D structure annotations {\bf (b)} and generate photo-realistic 2D images {\bf (c)}.}
\label{fig:teaser}
\end{figure}

When perceiving 3D scenes, humans are remarkably effective in using salient global structures such as lines, contours, planes, smooth surfaces, symmetries, and repetitive patterns. Thus, if a reconstruction algorithm can take advantage of such global information, it is natural to expect the algorithm to obtain more accurate results. Traditionally, however, it has been computationally challenging to reliably detect such global structures from noisy local image features. Recently, deep learning-based methods have shown promising results in detecting various forms of structure directly from the images, including lines~\cite{HuangWZDGM18,ZhouQZSCWM19}, planes~\cite{LiuYCYF18,YangZ18,LiuKGFK19,YuZLZG19}, cuboids~\cite{DwibediMBR16}, floorplans~\cite{LiuWKF17,LiuWF18}, room layouts~\cite{LeeBMR17,ZouCSH18,SunHSC19}, abstracted 3D shapes~\cite{TulsianiSGEM17,WuXLTTTF18}, and smooth surfaces~\cite{GroueixFKRA18}.

With the fast development of deep learning methods comes the need for large amounts of accurately annotated data. In order to train the proposed neural networks, most prior work collects their own sets of images and manually label the structure of interest in them. Such a strategy has several shortcomings. \emph{First}, due to the tedious process of manually labeling and verifying all the structure instances (\eg, line segments) in each image, existing datasets typically have limited sizes and scene diversity. And the annotations may also contain errors. \emph{Second}, since each study primarily focuses on one type of structure, none of these datasets has multiple types of structure labeled. As a result, existing methods are unable to exploit relations between different types of structure (\eg, lines and planes) as humans do for effective, efficient, and robust 3D reconstruction.

In this paper, we present a large synthetic dataset with rich annotations of 3D structure \emph{and} photo-realistic 2D renderings of indoor man-made environments (\figref{fig:teaser}). At the core of our dataset design is a unified representation of 3D structure which enables us to efficiently capture multiple types of 3D structure in the scene. Specifically, the proposed representation considers any structure as \emph{relationship} among \emph{geometric primitives}. For example, a ``wireframe'' structure encodes the incidence and intersection relationship between line segments, whereas a ``cuboid'' structure encodes the rotational and reflective symmetry relationship among its planar faces. With our ``primitive + relationship'' representation, one can easily derive the ground truth annotations for a wide variety of semi-global and global structures (\eg, lines, wireframes, planes, regular shapes, floorplans, and room layouts), and also exploit their relations in future data-driven approaches (\eg, the wireframe formed by intersecting planar surfaces in the scene).

\begin{table}[t]
\scriptsize
\renewcommand{\arraystretch}{1.2}
\centering
\caption{An overview of datasets with structure annotations. $^\dag$: The actual numbers are not explicitly given and hard to estimate, because these datasets contain images from Internet (LSUN Room Layout, PanoContext), or multiple sources (LayoutNet). $^*$: Dataset is unavailable online at the time of publication.}
\label{tab:dataset}
\begin{tabular}{l|cccc}
\hline
Datasets & \#Scenes & \#Rooms & \#Frames & Annotated structure \\
\hline
PlaneRCNN~\cite{LiuKGFK19}   & - & - & 100,000 & planes    \\
Wireframe~\cite{HuangWZDGM18} & - & - & 5,462 & wireframe (2D) \\
SceneCity 3D~\cite{ZhouQZSCWM19} & 230 & - & 23,000 & wireframe (3D) \\
SUN Primitive~\cite{XiaoRT12} & - & - & 785 & cuboids, other primitives \\
LSUN Room Layout~\cite{LSUN16} & - & n/a$^\dag$ & 5,394 & cuboid layout \\
PanoContext~\cite{ZhangSTX14} & - & n/a$^\dag$ & 500 (pano) & cuboid layout \\
LayoutNet~\cite{ZouCSH18} & - & n/a$^\dag$ & 1,071 (pano) & cuboid layout \\
MatterportLayout$^*$~\cite{ZouSPCSWCH19} & - & n/a$^\dag$ & 2,295 (RGB-D pano) & Manhattan layout \\
Raster-to-Vector~\cite{LiuWKF17} & 870 & - & - & floorplan \\
\hline
Structured3D & 3,500 & 21,835 & 196,515 & ``primitive + relationship'' \\
\hline
\end{tabular}
\end{table}

To create a large-scale dataset with the aim of facilitating research on data-driven methods for structured 3D scene understanding, we leverage the availability of professional interior designs and millions of production-level 3D object models -- all coming with fine geometric details and high-resolution textures (\figref{fig:teaser}(a)). We first use computer programs to automatically extract information about 3D structure from the original house design files. As shown in \figref{fig:teaser}(b), our dataset contains rich annotations of 3D room structure including a variety of geometric primitives and relationships. To further generate photo-realistic 2D images (\figref{fig:teaser}(c)), we utilize industry-leading rendering engines to model the lighting conditions. Currently, our dataset consists of more than 196k images of 21,835 rooms in 3,500 scenes (\ie, houses).

To showcase the usefulness and uniqueness of the proposed Structured3D dataset, we train deep networks for room layout estimation on a subset of the dataset. We show that the models trained on both synthetic and real data outperform the models trained on real data only. Further, following the spirit of~\cite{TsaiHSSYC18,ChenLCVG19}, we show how multi-modal annotations in our dataset can benefit domain adaptation tasks.

In summary, the {\bf main contributions} of this paper are:
\begin{itemize}
\item We create the Structured3D dataset, which contains rich ground truth 3D structure annotations of 21,835 rooms in 3,500 scenes, and more than 196k photo-realistic 2D renderings of the rooms.

\item We introduce a unified ``primitive + relationship'' representation. This representation enables us to efficiently capture a wide variety of semi-global or global 3D structures and their mutual relationships.

\item We verify the usefulness of our dataset by using it to train deep networks for room layout estimation and demonstrating improved performance on public benchmarks.
\end{itemize}

\begin{figure}[t]
\scriptsize
\centering
\begin{tabular}{cccc}
\includegraphics[height=0.88in]{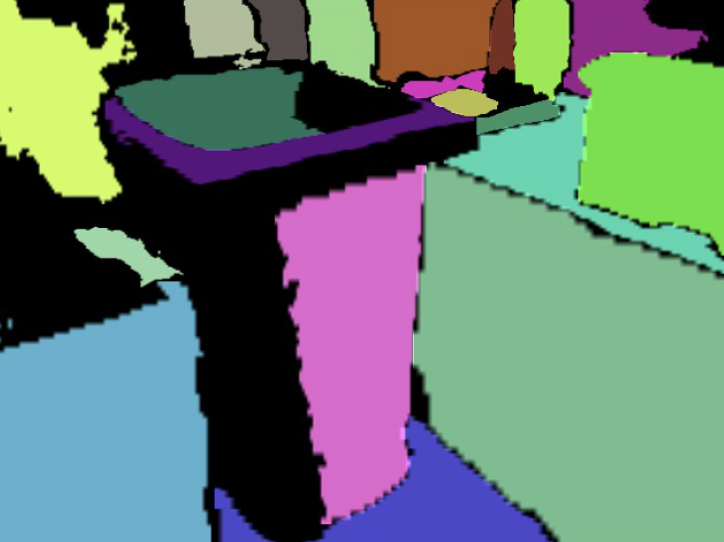} &
\includegraphics[height=0.88in]{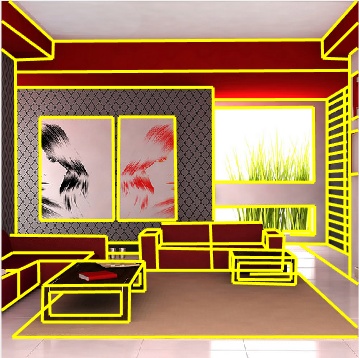} &
\includegraphics[height=0.88in]{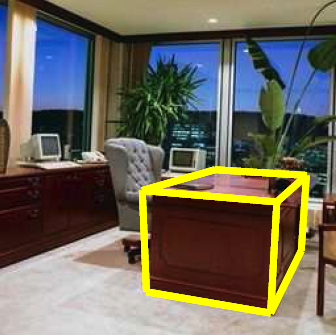} & 
\includegraphics[height=0.88in]{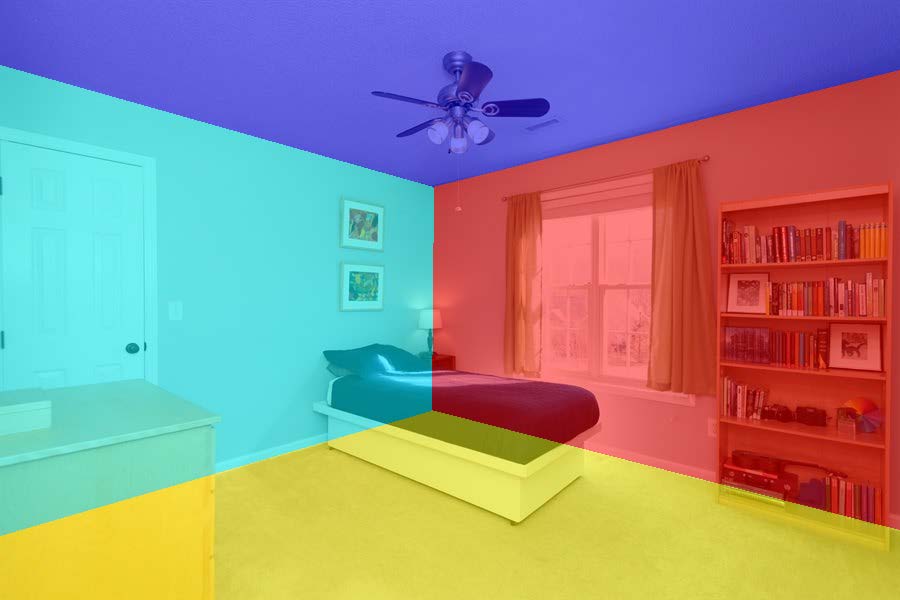} \\
(a) Plane~\cite{LiuKGFK19} & (b) Wireframe~\cite{HuangWZDGM18} & (c) Cuboid~\cite{DwibediMBR16} & (d) Room layout~\cite{LSUN16} \\
\end{tabular}
\begin{tabular}{cc}
\includegraphics[height=0.86in]{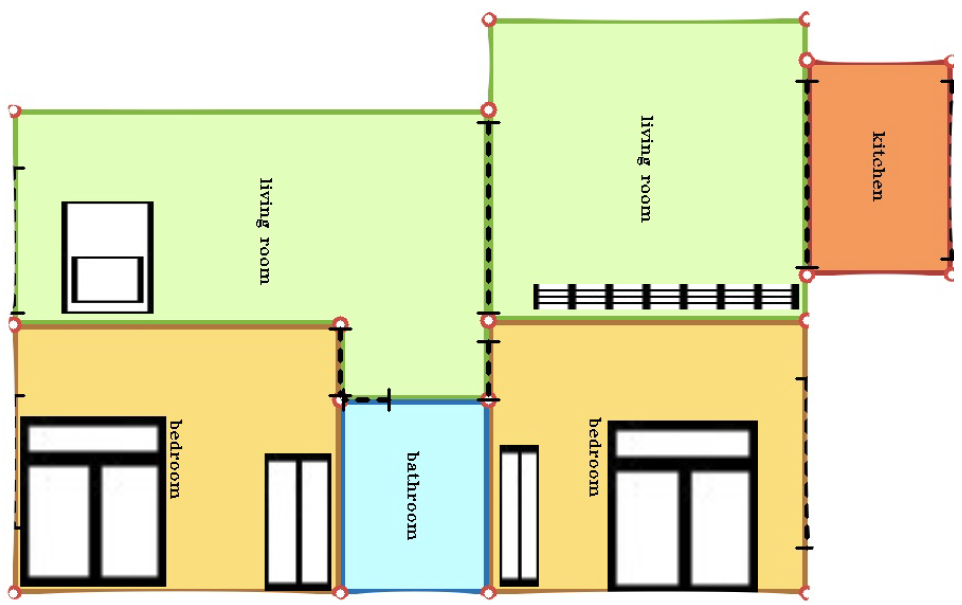} & 
\fbox{\includegraphics[height=0.86in]{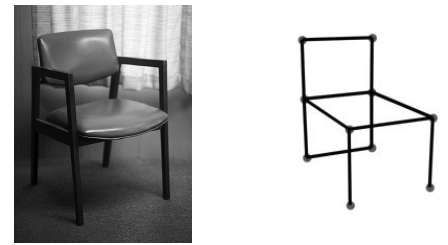}}
\fbox{\includegraphics[height=0.86in]{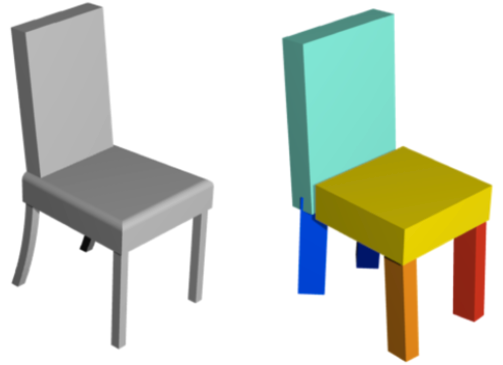}} \\
(e) Floorplan~\cite{LiuWKF17} & (f) Abstracted 3D shape (wireframe~\cite{WuXLTTTF18} and cuboid~\cite{TulsianiSGEM17}) \\
\end{tabular}
\caption{Example annotations of structure in existing datasets. The reference number indicates the paper from which the illustration is originally from.}
\label{fig:data}
\end{figure}

\section{Related Work}

\noindent{\bf Datasets.} \tabref{tab:dataset} summarizes existing datasets for structured 3D scene modeling. Additionally, \cite{TulsianiSGEM17,WuXLTTTF18} provide datasets with structured representations of single objects. We show example annotations in these datasets in \figref{fig:data}. Note that ground truth annotations in most datasets are manually labeled. This is one main reason why all these datasets have limited size, \ie, contain no more than a few thousand images. One exception is \cite{LiuKGFK19}, which employs a multi-model fitting algorithm to automatically extract planes from 3D scans in the ScanNet dataset~\cite{DaiCSHFN17}. But such algorithms are sensitive to data noises and outliers, thus introduce errors in the annotations (\figref{fig:data}(a)). Similar to our work, SceneCity 3D~\cite{ZhouQZSCWM19} also contains synthetic images with ground truth automatically extracted from CAD models. But the number of scenes is limited to 230. Further, none of these datasets has more than one type of structure labeled, although different types of structure often have strong relations among them. For example, from the wireframe in \figref{fig:data}(b) humans can easily identify other types of structure such as planes and cuboids. Our new dataset sets to bridge the gap between what is needed to train machine learning models to achieve human-level holistic 3D scene understanding and what is being offered by existing datasets.

Note that our dataset is very different from other popular large-scale 3D datasets, such as NYU v2~\cite{SilbermanHKF12}, SUN RGB-D~\cite{SongLX15}, 2D-3D-S~\cite{ArmeniSZJBFS16,ArmeniSZS17}, ScanNet~\cite{DaiCSHFN17}, and Matterport3D~\cite{ChangDFHNSSZZ17}, in which the ground truth 3D information is stored in the format of point clouds or meshes. These datasets lack ground truth annotations of semi-global or global structures. While it is theoretically possible to extract 3D structure by applying structure detection algorithms to the point clouds or meshes (\eg, extracting planes from ScanNet as did in~\cite{LiuKGFK19}), the detection results are often noisy and even contain errors. In addition, for some types of structure like wireframes and room layouts, how to reliably detect them from raw sensor data remains an active research topic in computer vision.

In recent years, synthetic datasets have played an important role in the successful training of deep neural networks. Notable examples for indoor scene understanding include SUNCG~\cite{SongYZCSF17}, SceneNet RGB-D~\cite{McCormacHLD17}, and InteriorNet~\cite{LiSMCTYHTL18}. These datasets exceed real datasets in terms of scene diversity and frame numbers. But just like their real counterparts, these datasets lack ground truth structure annotations. Another issue with some synthetic datasets is the degree of realism in both the 3D models and the 2D renderings. \cite{ZhangSYSLJF17} shows that physically-based rendering could boost the performance of various indoor scene understanding tasks. To ensure the quality of our dataset, we make use of 3D room models created by professional designers and the state-of-the-art industrial rendering engines. \tabref{tab:3d_dataset} summarizes the differences of 3D scene datasets.

\smallskip
\noindent{\bf Room layout estimation.} Room layout estimation aims to reconstruct the enclosing structure of the indoor scene, consisting of walls, floor, and ceiling. Existing public datasets (\eg, PanoContext~\cite{ZhangSTX14} and LayoutNet~\cite{ZouCSH18}) assume a simple box-shaped layout. PanoContext~\cite{ZhangSTX14} collects about 500 panoramas from the SUN360 dataset~\cite{XiaoEOT12}, LayoutNet~\cite{ZouCSH18} extends the layout annotations to include panoramas from 2D-3D-S~\cite{ArmeniSZS17}. Recently, MatterportLayout~\cite{ZouSPCSWCH19} collects 2,295 RGB-D panoramas from Matterport3D~\cite{ChangDFHNSSZZ17} and extends annotations to Manhattan layout. We note that all room layout in these real datasets is manually labeled by the human. Since the room structure may be occluded by furniture and other objects, the ``ground truth'' inferred by humans may not be consistent with the actual layout. In our dataset, all ground truth 3D annotations are automatically extracted from the original house design files.

\begin{table}[t]
\scriptsize
\renewcommand{\arraystretch}{1.2}
\setlength{\tabcolsep}{4pt}
\centering
\caption{Comparison of 3D scene datasets. $^\dag$: Meshes are obtained by 3D reconstruction algorithm. Notations for applications: O (object detection), U (scene understanding), S (image synthesis), M (structured 3D modeling).}
\label{tab:3d_dataset}
\begin{tabular}{l|cccc}
\hline
Datasets & Scene design type & 3D annotation & 2D rendering & Applications \\
\hline
NYU v2~\cite{SilbermanHKF12} & Real & Raw RGB-D & Real images & O U \\
SUN RGB-D~\cite{SongLX15} & Real & Raw RGB-D & Real images & O U \\
2D-3D-S~\cite{ArmeniSZJBFS16,ArmeniSZS17} & Real & Mesh$^\dag$ & Real images & O U \\
ScanNet~\cite{DaiCSHFN17} & Real & Mesh$^\dag$ & Real images & O U \\
Matterport3D~\cite{ChangDFHNSSZZ17} & Real & Mesh$^\dag$ & Real images & O U \\
\hline
SUNCG~\cite{SongYZCSF17} & Amateur & Mesh & n/a & O U \\
SceneNet RGB-D~\cite{McCormacHLD17} & Random & Mesh & Photo-realistic & O U \\
InteriorNet~\cite{LiSMCTYHTL18} & Professional & n/a & Photo-realistic & O U S \\
\hline
Structured3D & Professional & 3D structures & Photo-realistic & O U S M \\
\hline
\end{tabular}
\end{table}

\section{A Unified Representation of 3D Structure}

The main goal of our dataset is to provide rich annotations of ground truth 3D structure. A naive way to do so is generating and storing different types of 3D annotations in the same format as existing works, like wireframes as in~\cite{HuangWZDGM18}, planes as in~\cite{LiuKGFK19}, floorplans as in~\cite{LiuWKF17}, and so on. But this leads to a lot of redundancy. For example, planes in man-made environments are often bounded by a number of line segments, which are part of the wireframe. Even worse, by representing wireframes and planes separately, the relationships between them are lost. In this paper, we present a unified representation in order to minimize redundancy while preserving mutual relationships. We show how the most common types of structure studied in the literature (\eg, planes, cuboids, wireframes, room layouts, and floorplans) can be derived from our representation.

\begin{figure}[t]
\scriptsize
\centering
\begin{tabular}{ccc}
\includegraphics[width=0.3\linewidth]{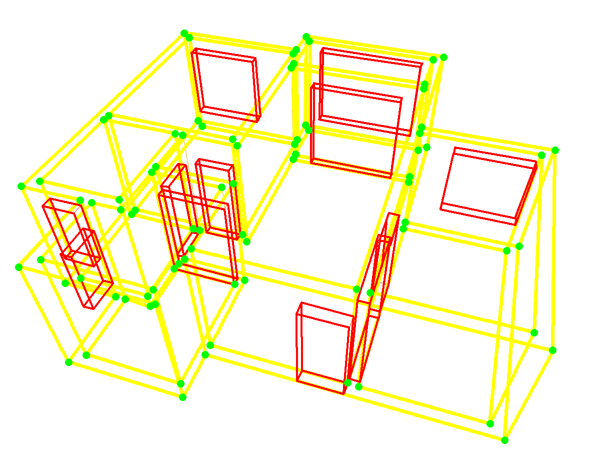} &
\includegraphics[width=0.3\linewidth]{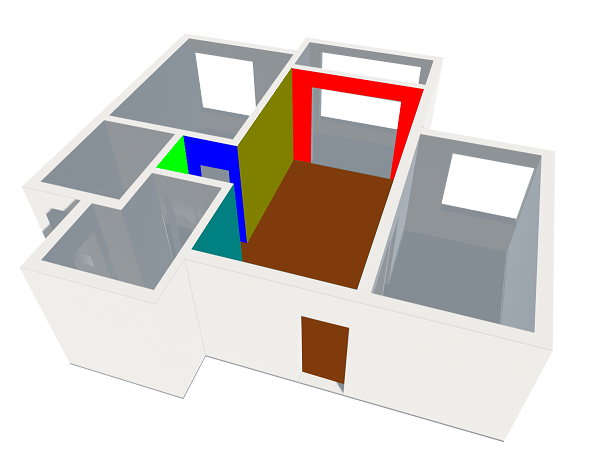} &
\includegraphics[width=0.3\linewidth]{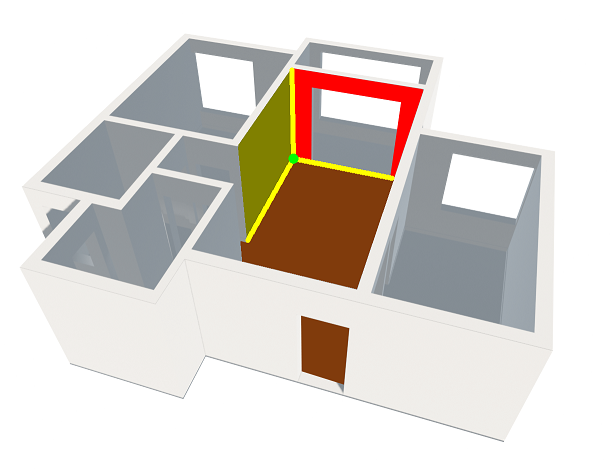} \\
(a) Primitives: junctions and lines & (b) Primitives: planes & (c) Relationships: $R_1$ and $R_2$ \\
\includegraphics[width=0.3\linewidth]{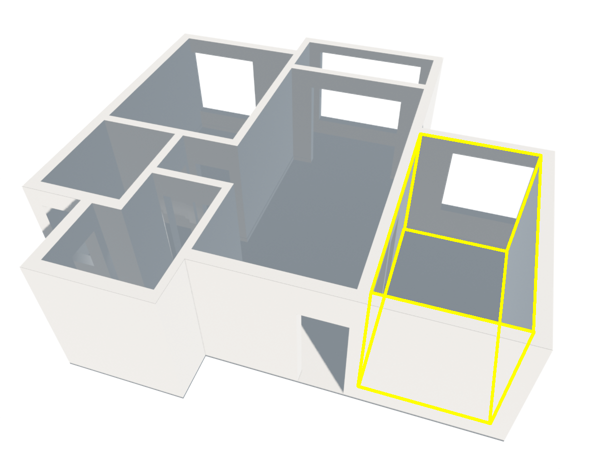} &
\includegraphics[width=0.3\linewidth]{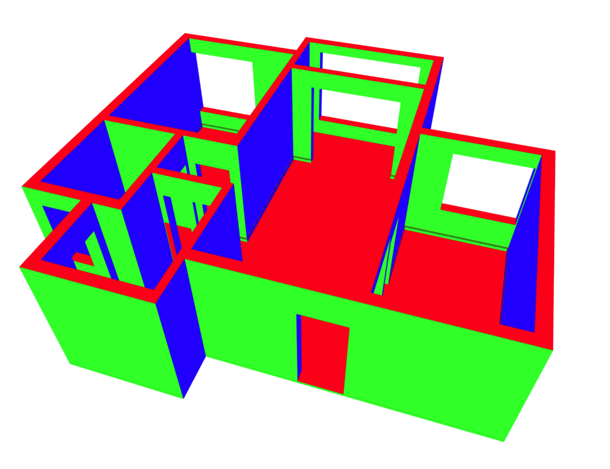} &
\includegraphics[width=0.3\linewidth]{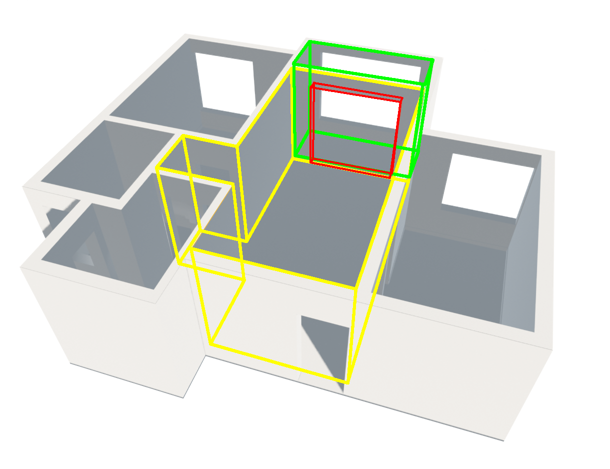} \\
(d) Relationships: $R_3$ & (e) Relationships: $R_4$ & (f) Relationships: $R_5$
\end{tabular}
\caption{The ground truth 3D structure annotations in our dataset are represented by primitives and relationships. {\bf (a)}: Junctions and lines. {\bf (b)}: Planes. We highlight the planes in a single room. {\bf (c)}: Plane-line and line-junction relationships. We highlight a junction, the three lines intersecting at the junction, and the planes intersecting at each of the lines. {\bf (d)}: Cuboids. We highlight one cuboid instance. {\bf (e)}: Manhattan world. We use different colors to denote planes aligned with different directions. {\bf (f)}: Semantic objects. We highlight a ``room'', a ``balcony'', and the ``door'' connecting them.}
\label{fig:3Dgt}
\end{figure}

Our representation of the structure is largely inspired by the early work of Witkin and Tenenbaum~\cite{Witkin1983}, which characterizes structure as \emph{``a shape, pattern, or configuration that replicates or continues with little or no change over an interval of space and time''}. Accordingly, to describe any structure, we need to specify: (i) what pattern is continuing or replicating (\eg, a patch, an edge, or a texture descriptor), and (ii) the domain of its replication or continuation. In this paper, we call the former {\bf primitives} and the latter {\bf relationships}.

\subsection{The ``Primitive + Relationship'' Representation}

We now show how to describe a man-made environment using a unified representation. For ease of exposition, we assume all objects in the scene can be modeled by piece-wise planar surfaces. But our representation can be easily extended to more general surfaces. An illustration of our representation is shown in \figref{fig:3Dgt}.

\smallskip
\noindent{\bf Primitives.} Generally, a man-made scene has the following geometric primitives:
\begin{itemize}
\item {\bf Planes $\P$}: We model the scene as a collection of planes $\P = \{p_1, p_2, \ldots \}$. Each plane is described by its parameters $p = \{\n, d\}$, where $\n$ and $d$ denote the surface normal and the distance to the origin, respectively.
\item {\bf Lines $\L$}: When two planes intersect in the 3D space, a line is created. We use $\L = \{l_1, l_2, \ldots \}$ to represent the set of all 3D lines in the scene.
\item {\bf Junction points $\X$}: When two lines meet in the 3D space, a junction point is formed. We use $\X = \{x_1, x_2, \ldots \}$ to represent the set of all junction points.
\end{itemize}

\smallskip
\noindent{\bf Relationships.} Next, we define some common types of relationships between the geometric primitives:
\begin{itemize}
\item {\bf Plane-line relationships ($R_1$)}: We use a matrix $W_1$ to record all incidence and intersection relationships between planes in $\P$ and lines in $\L$. Specifically, the $ij$-th entry of $W_1$ is 1 if $l_i$ is on $p_j$, and 0 otherwise. Note that two planes are intersected at some line if and only if the corresponding entry in $W_1^TW_1$ is nonzero.

\item {\bf Line-point relationships ($R_2$)}: Similarly, we use a matrix $W_2$ to record all incidence and intersection relationships between lines in $\L$ and points in $\X$. Specifically, the $mn$-th entry of $W_2$ is 1 if $x_m$ is on $l_n$, and 0 otherwise. Note that two lines are intersected at some junction if and only if the corresponding entry in $W_2^TW_2$ is nonzero.

\item {\bf Cuboids ($R_3$)}: A cuboid is a special arrangement of plane primitives with rotational and reflection symmetry along x-, y- and z-axes. The corresponding symmetry group is the dihedral group $D_{2h}$.

\item {\bf Manhattan world ($R_4$)}: This is a special type of 3D structure commonly used for indoor and outdoor scene modeling. It can be viewed as a \emph{grouping} relationship, in which all the plane primitives can be grouped into three classes, $\P_1$, $\P_2$, and $\P_3$, $\P = \bigcup_{i=1}^3 \P_i$. Further, each class is represented by a single normal vector $\n_i$, such that $\n_i^T \n_j = 0, i\neq j$.

\item {\bf Semantic objects ($R_5$)}: Semantic information is critical for many 3D computer vision tasks. It can be regarded as another type of \emph{grouping} relationship, in which each semantic object instance corresponds to one or more primitives defined above. For example, each ``wall'', ``ceiling'', or ``floor'' instance is associated with one plane primitive; each ``chair'' instance is associated with a set of multiple plane primitives. Further, such a grouping is hierarchical. For example, we can further group one floor, one ceiling, and multiple walls to form a ``living room'' instance. And a ``door'' or a ``window'' is an opening which connects two rooms (or one room and the outer space).

\end{itemize}

Note that the relationships are not mutually exclusive, in the sense that a primitive can belong to multiple relationship instances of the same type or different types. For example, a plane primitive can be shared by two cuboids, and at the same time belong to one of the three classes in the Manhattan world model.

\smallskip
\noindent{\bf Discussion.} The primitives and relationships we discussed above are just a few most common examples. They are by no means exhaustive. For example, our representation can be easily extended to include other primitives such as parametric surfaces. And besides cuboids, there are many other types of regular or symmetric shapes in man-made environments, where type corresponds to a different symmetry group.

Our representation of 3D structures is also related to the graph representations in semantic scene understanding~\cite{HuangQZXXZ18,ArmeniHGZFMS19,WangLWSCR19}.
As these graphs focus on semantics, geometry is represented in simplified manners by (i) 6D object poses and (ii) coarse, discrete spatial relations such as "supported by", "front", "back", and "adjacent". In contrast, our representation focuses on modeling the scene geometry using fine-grained primitives (\ie, junctions, lines, and planes) and relationships (in terms of topology and regularities). Thus, it is highly complementary to the scene graphs in prior work. Intuitively, it can be used for geometric analysis and synthesis tasks, in a similar way as scene graphs are used for semantic scene understanding.

\subsection{Relation to Existing Models}

Given our representation which contains primitives $\mathcal{P} = \{\P, \L, \X\}$ and relationships $\mathcal{R} = \{R_1, R_2, \ldots\}$, we show how several types of 3D structure commonly studied in the literature can be derived from it. We again refer readers to \figref{fig:data} for illustrations of these structures.

\smallskip
\noindent{\bf Planes}: A large volume of studies in the literature model the scene as a collection of 3D planes, where each plane is represented by its parameters and boundary. To generate such a model, we simply use the plane primitives $\P$. For each $p\in \P$, we further obtain its boundary by using  matrix $W_1$ in $R_1$ to find all the lines in $\L$ that form an incidence relationship with $p$.

\smallskip
\noindent{\bf Wireframes}: A wireframe consists of lines $\L$ and junction points $\P$, and their incidence and intersection relationships ($R_2$).

\smallskip
\noindent{\bf Cuboids}: This model is same as $R_3$.

\smallskip
\noindent{\bf Manhattan layouts}: A Manhattan room layout model includes a ``room'' as defined in $R_5$ which also satisfies the Manhattan world assumption ($R_4$).

\smallskip
\noindent{\bf Floorplans}: A floorplan is a 2D vector representation that consists of a set of line segments and semantic labels (\eg, room types). To obtain such a vector representation, we can identify all lines in $\L$ and junction points in $\X$ which lie on a ``floor'' (as defined in $R_5$). To further obtain the semantic room labels, we can project all ``rooms'', ``doors'', and ``windows'' (as defined in $R_5$) to this floor.

\smallskip
\noindent{\bf Abstracted 3D shapes}: In addition to room structures, our representation can also be applied to individual 3D object models to create abstractions in the form of wireframes or cuboids, as described above.

\section{The Structured3D Dataset}

\begin{figure}[t]
\scriptsize
\centering
\begin{tabular}{c|c}
\includegraphics[height=1.2in]{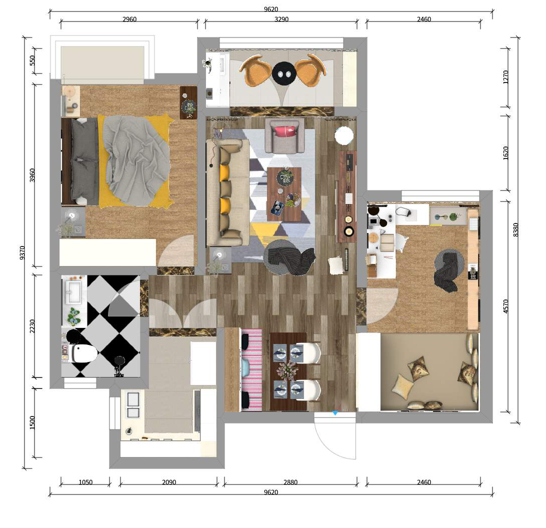} 
\includegraphics[height=1.2in]{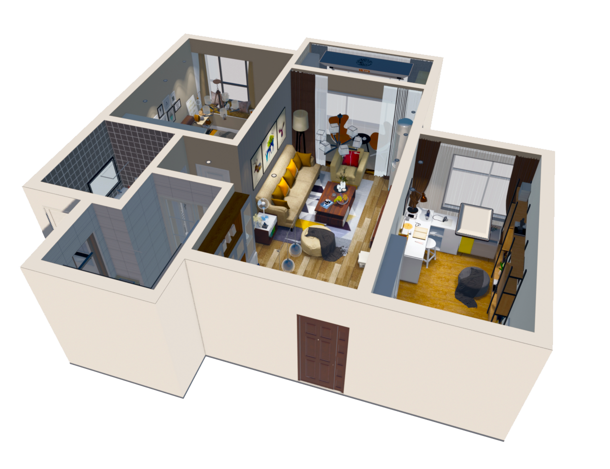} &
\includegraphics[height=1.2in]{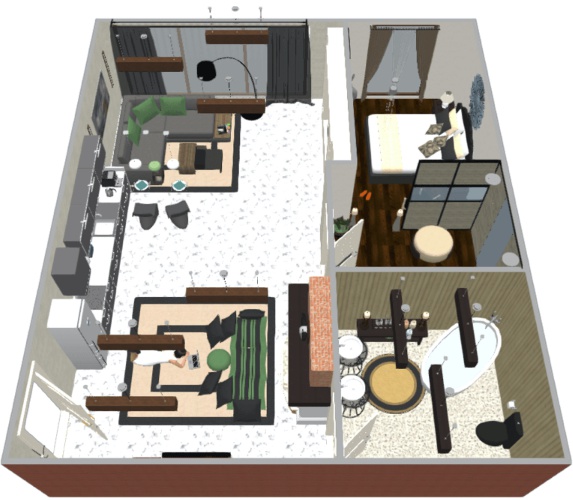} \\
\multicolumn{1}{c}{(a)} & (b)
\end{tabular}
\caption{Comparison of 3D house designs. {\bf (a)}: The 3D models in our database are created by professional designers using high-quality furniture models from world-leading manufacturers. Most designs are being used in real-world production. {\bf (b)}: The 3D models in SUNCG dataset~\cite{SongYZCSF17} are created using Planner 5D~\cite{planner}, an online tool for amateur interior design.}
\label{fig:design}
\end{figure}

Our unified representation enables us to encode a rich set of geometric primitives and relationships for structured 3D modeling. With this representation, our ultimate goal is to build a dataset that can be used to train machines to achieve the human-level understanding of the 3D environment.

As a first step towards this goal, in this section, we describe our ongoing effort to create a large-scale dataset of indoor scenes which include (i) ground truth 3D structure annotations of the scene and (ii) realistic 2D renderings of the scene. Note that in this work we focus on extracting ground truth annotations on the room structure only. We plan to extend our dataset to include 3D structure annotations of individual furniture models in the future.

In the following, we describe our general procedure to create the dataset. We refer readers to the supplementary materials for additional details, including dataset statistics and example annotations.

\subsection{Extraction of Structured 3D Models}

To extract a ``primitive + relationship'' scene representation, we utilize a large database of house designs hand-crafted by professional designers. An example design is shown in \figref{fig:design}(a). All information of the design is stored in an industry-standard format in the database so that specifications about the geometry (\eg, the precise size of each wall), textures and materials, and functions (\eg, which room the wall belongs to) of all objects can be easily retrieved.

From the database, we have selected 3,500 house designs with 21,835 rooms. We created a computer program to automatically extract all the geometric primitives associated with the room structure, which consists of the ceiling, floor, walls, and openings (doors and windows). Given the precise measurements and associated information of these entities, it is straightforward to generate all planes, lines, and junctions, as well as their relationships ($R_1$ and $R_2$).

Since the measurements are highly accurate and noise-free, other types of relationship such a Manhattan world ($R_3$) and cuboids ($R_4$) can also be easily obtained by clustering the primitives, followed by a geometric verification process. Finally, to include semantic information ($R_5$) into our representation, we map the relevant labels provided by the professional designers to the geometric primitives in our representation. \figref{fig:3Dgt} shows examples of the extracted geometric primitives and relationships.

\begin{figure}[t]
\scriptsize
\centering
\begin{tabular}{ccc}
\includegraphics[width=0.32\linewidth]{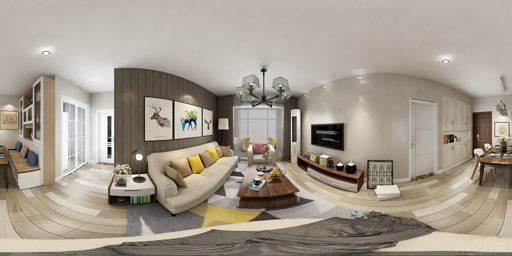} &
\includegraphics[width=0.32\linewidth]{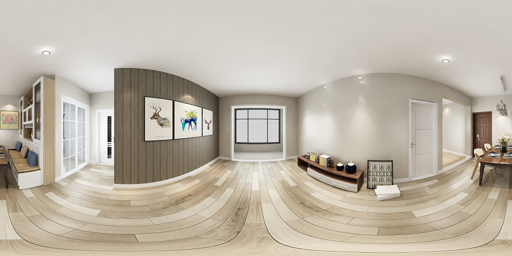} &
\includegraphics[width=0.32\linewidth]{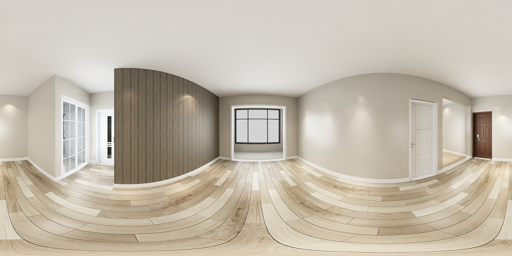} \\
(a) Original & (b) Simple configuration & (c) Empty configuration \\
\includegraphics[width=0.32\linewidth]{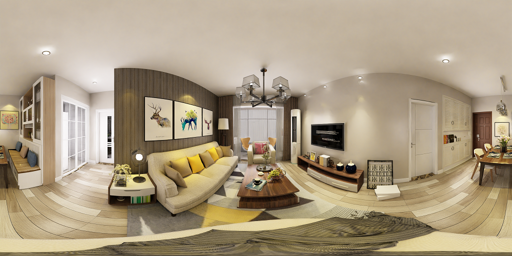} &
\includegraphics[width=0.32\linewidth]{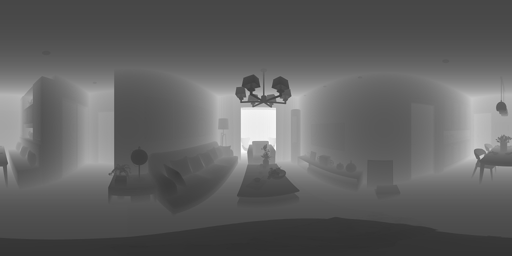} &
\includegraphics[width=0.32\linewidth]{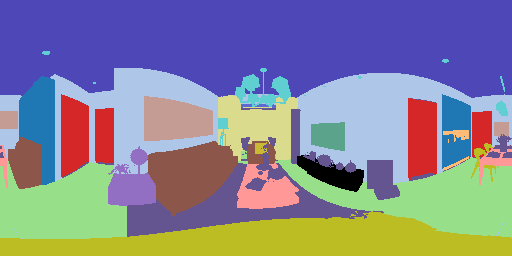} \\
(d) Lighting & (e) Depth & (f) Semantic labels
\end{tabular}
\caption{Examples of our rendered panoramic images.}
\label{fig:rendering}
\end{figure}

\subsection{Photo-realistic 2D Rendering}

To ensure the quality of our 2D renderings, our rendering engine is developed in collaboration with a company specialized in interior design rendering. Our engine uses a well-known ray-tracing method~\cite{PurcellBMH02}, a Monte Carlo approach to approximating realistic Global Illumination (GI), for RGB rendering. The other ground truth images are obtained by a customized path-tracer renderer on top of Intel Embree~\cite{WaldWBJE14}, an open-source collection of ray-tracing kernels for x86 CPUs.

Each room is manually created by professional designers with over one million CAD models of furniture from world-leading manufacturers. These high-resolution furniture models are measured in real-world dimensions and being used in real production. A default lighting setup is also provided. \figref{fig:design} compares the 3D models in our database with those in SUNCG~\cite{SongYZCSF17}, which are created using Planner 5D~\cite{planner}, an online tool for amateur interior design. 

\begin{figure}[t]
\centering
\begin{tabular}{cc|cc}
\includegraphics[height=0.8in,width=0.24\linewidth]{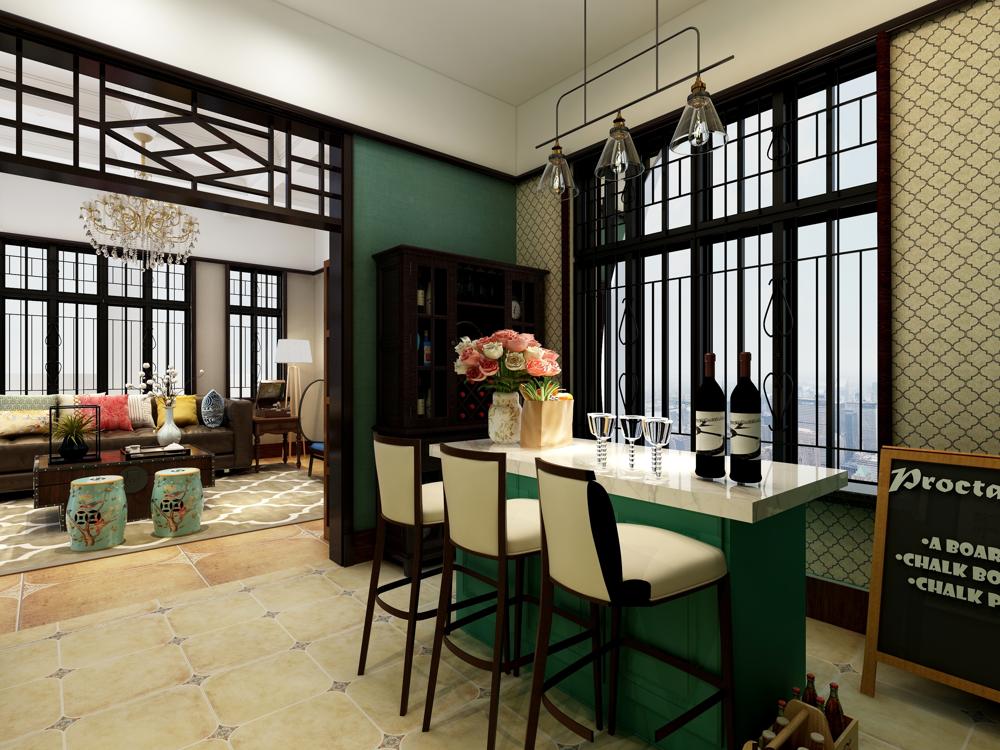} &
\includegraphics[height=0.8in,width=0.24\linewidth]{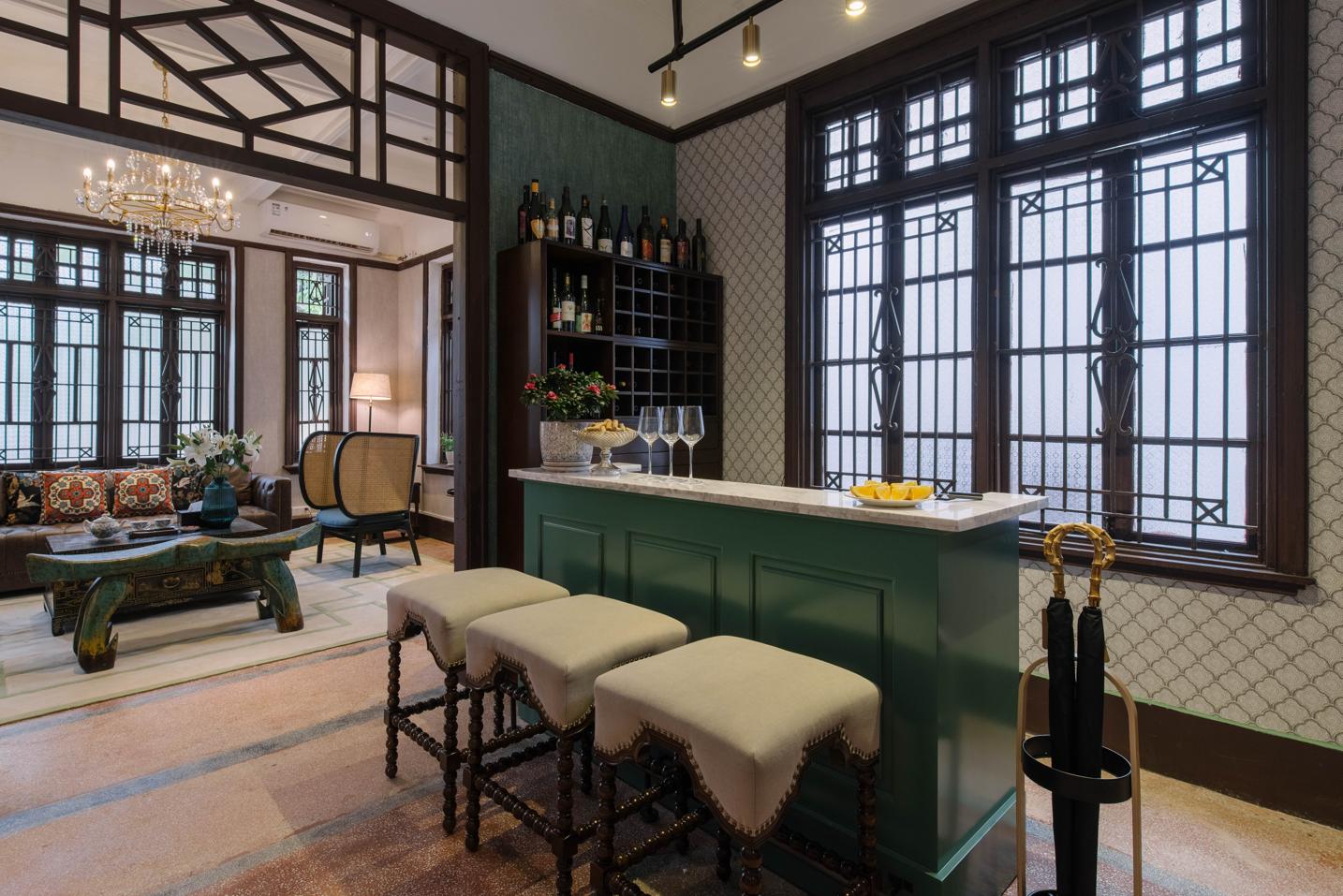} &
\includegraphics[height=0.8in,width=0.24\linewidth]{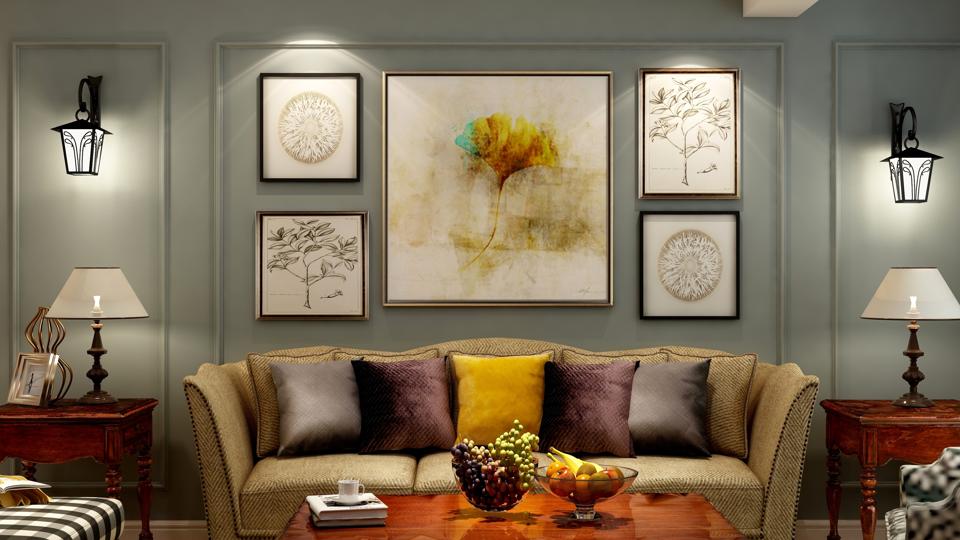} &
\includegraphics[height=0.8in,width=0.24\linewidth]{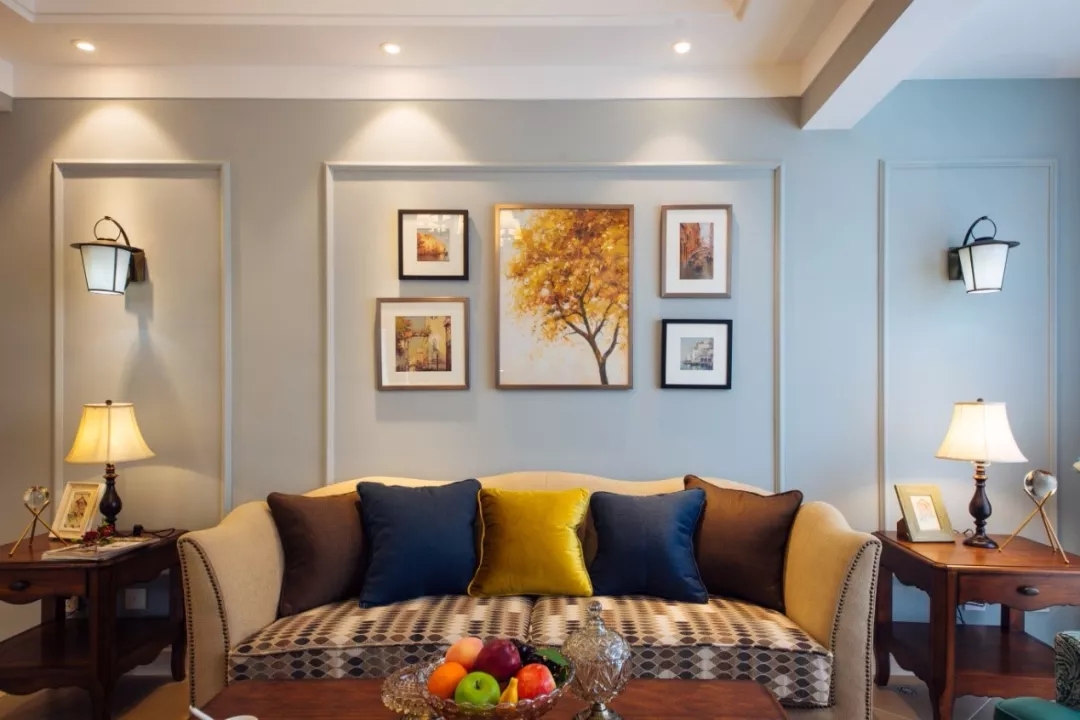}
\end{tabular}
\caption{Photo-realistic rendering vs. real-world decoration. The first and third columns are rendered images.
}
\label{fig:rendering_vs_real}
\end{figure}

At the time of rendering, a panoramic or pin-hole camera is placed at random locations not occupied by objects in the room. We use $512 \times 1024$ resolution for panoramas and $720 \times 1280$ for perspective images. \figref{fig:rendering} shows example panoramas rendered by our engine. For each room, we generate different configurations (full, simple, and empty) by removing some or all the furniture. We also modify the lighting setup to generate images with different temperatures. For each image, our dataset also includes the depth map and semantic mask. \figref{fig:rendering_vs_real} illustrates the degree of photo-realism of our dataset, where we compare the rendered images with photos of real decoration guided by the design.

\subsection{Use Cases}

Due to the unique characteristics of our dataset, we envision it contributing to computer vision research in terms of both methodology and applications.

\smallskip
\noindent{\bf Methodology.} As our dataset contains multiple types of 3D structure annotations as well as ground truth labels (\eg, semantic maps, depth maps, and 3D object bounding boxes), it enables researchers to design novel multi-modal or multi-task approaches for a variety of vision tasks. As an example, we show in Section~\ref{sec:exp} that, by leveraging multi-modal annotations in our dataset, we can boost the performance of existing room layout estimation methods in the domain adaptation framework.

\smallskip
\noindent{\bf Applications.} Our dataset also facilitates research on a number of problems and applications. For example, as shown in \tabref{tab:dataset}, all publicly available datasets for room layout estimation are limited to simple cuboid rooms. Our dataset is the first to provide the general (non-cuboid) room layout annotations. As another example, existing datasets for floorplan reconstruction~\cite{LiuWF18,ChenJ19} contain about 100-150 scenes, whereas our dataset includes 3,500 scenes.

Another major line of research that would benefit from our dataset is image synthesis. With a photo-realistic rendering engine, we are able to generate images given \emph{any} scene configurations and viewpoints. These images may be used as ground truth for tasks including image inpainting (\eg, completing an image when certain furniture is removed) and novel view synthesis.

\smallskip
Finally, we would like to emphasize the potential of our dataset in terms of extension capabilities. As we mentioned before, the unified representation enables us to include many other types of structure in the dataset. As for 2D rendering, depending on the application, we can easily simulate different effects such as lighting conditions, fisheye and novel camera designs, motion blur, and imaging noise. Furthermore, the dataset may be extended to include videos for applications such as visual SLAM~\cite{CadenaCCLSNRL16}.

\section{Experiments}
\label{sec:exp}

\subsection{Experiment Setup}

To demonstrate the benefits of our dataset, we use it to train deep neural networks for room layout estimation, an important task in structured 3D modeling.

\smallskip
\noindent{\bf Real dataset.} We use the same dataset as LayoutNet~\cite{ZouCSH18}. The dataset consists of images from PanoContext~\cite{ZhangSTX14} and 2D-3D-S~\cite{ArmeniSZS17}, including 818 training images, 79 validation images, and 166 test images. Note that both datasets only provide cuboid layout annotations.

\begin{table}[t]
\scriptsize
\renewcommand{\arraystretch}{1.2}
\setlength{\tabcolsep}{6pt}
\centering
\caption{Room layout statistics. $^\dag$: MatterportLayout is the only other dataset with non-cuboid layout annotations, but is unavailable at the time of publication.}
\label{tab:statistic}
\begin{tabular}{l|cccccccc}
\hline
\#Corners & 4 & 5 & 6 & 7 & 8 & 9 & 10+ & Total \\
\hline
MatterportLayout$^\dag$ & 1211 & 0 & 501 & 0 & 309 & 0 & 274 & 2295 \\
Structured3D & 13743 & 52 & 3727 & 30 & 1575 & 17 & 2691 & 21835 \\
\hline
\end{tabular}
\end{table}

\smallskip
\noindent{\bf Our Structured3D dataset.} In this experiment, we use a subset of panoramas with the original lighting and full configuration. Each panorama corresponds to a different room in our dataset. We show statistics of different room layouts in our dataset in \tabref{tab:statistic}. Since the current real dataset only contains cuboid layout annotations (\ie, 4 corners), we choose 12k panoramic images with the cuboid layout in our dataset. We split the images into 10k for training, 1k for validation, and 1k for testing.

\smallskip
\noindent{\bf Evaluation metrics.} Following~\cite{ZouCSH18,SunHSC19}, we adopt three standard metrics: (i) 3D IoU: intersection over union between predicted 3D layout and the ground truth, (ii) Corner Error (CE): normalized $\ell_2$ distance between predicted corner and ground truth, and (iii) Pixel Error (PE): pixel-wise error between predicted plane classes and ground truth.

\smallskip
\noindent{\bf Baselines.} We choose two recent CNN-based approaches, LayoutNet~\cite{ZouCSH18,ZouSPCSWCH19}\footnote[1]{\url{https://github.com/zouchuhang/LayoutNetv2}} and HorizonNet~\cite{SunHSC19}\footnote[2]{\url{https://github.com/sunset1995/HorizonNet}}, based on their performance and source code availability. LayoutNet uses a CNN to predict a corner probability map and a boundary map from the panorama and vanishing lines, then optimizes the layout parameters based on network predictions. HorizonNet represents room layout as three 1D vectors, \ie, boundary positions of floor-wall, and ceiling-wall, and the existence of wall-wall boundary. It trains CNNs to directly predict the three 1D vectors. In this paper, we follow the default training setting of the respective methods. For specific training procedures, please refer to the supplementary materials.

\subsection{Experiment Results}

\begin{table}[t]
\scriptsize
\renewcommand{\arraystretch}{1.2}
\centering
\caption{Quantitative evaluation under different training schemes. The best and the second best results are boldfaced and underlined, respectively.}
\label{tab:result:synthetic}
\begin{tabular}{l|c|ccc|ccc}
\hline
\multirow{2}{*}{Methods} & \multirow{2}{*}{Config.} & \multicolumn{3}{c|}{PanoContext} & \multicolumn{3}{c}{2D-3D-S}\\ \cline{3-8}
& & {3D IoU (\%) $\uparrow$} & {CE (\%) $\downarrow$} & {PE (\%) $\downarrow$} & {3D IoU (\%) $\uparrow$} & {CE (\%) $\downarrow$} & {PE (\%) $\downarrow$} \\
\hline
\multirow{4}{*}{LayoutNet~\cite{ZouCSH18,ZouSPCSWCH19}}
& s & 75.64 & 1.31 & 4.10 & 57.18 & 2.28 & 7.55 \\
& r & 84.15 & 0.64 & \underline{1.80} & 83.39 & 0.74 & 2.39 \\
& s + r & {\bf 84.96} & {\bf 0.61} & {\bf 1.75} & \underline{83.66} & \underline{0.71} & \underline{2.31} \\
& s $\rightarrow$ r & \underline{84.77} & \underline{0.63} & {1.89} & {\bf 84.04} & {\bf 0.66} & {\bf 2.08} \\
\hline
\multirow{4}{*}{HorizonNet~\cite{SunHSC19}}
& s & 75.89 & 1.13 & 3.15 & 67.66 & 1.18 & 3.94 \\
& r & 83.42 & 0.73 & 2.09 & 84.33 & 0.64 & 2.04 \\
& s + r &  \underline{84.45} & \underline{0.70} & \underline{1.89} & \underline{84.36} & {\bf 0.59} & \underline{1.90} \\
& s $\rightarrow$ r & {\bf 85.27} & {\bf 0.66} & {\bf 1.86} & {\bf 86.01} & \underline{0.61} & {\bf 1.84} \\
\hline
\end{tabular}
\end{table}

\noindent{\bf Augmenting real datasets.} In this experiment, we train LayoutNet and HorizonNet in four different manners: (i) training only on our synthetic dataset (``{\bf s}''), (ii) training only on the real dataset (``{\bf r}''), (iii) training on the synthetic and real dataset with Balanced Gradient Contribution (BGC)~\cite{RosSAW16} (``{\bf s + r}''), and (iv) pre-training on our synthetic dataset, then fine-tuning on the real dataset (``{\bf s $\rightarrow$ r}'').  We adopt the training set of LayoutNet as the real dataset in this experiment. The results are shown in \tabref{tab:result:synthetic}. As one can see, augmenting real datasets with our synthetic data boosts the performance of both networks. We refer readers to supplementary materials for more qualitative results.

\begin{table}[t]
\scriptsize
\renewcommand{\arraystretch}{1.2}
\centering
\caption{Quantitative evaluation using varying synthetic data size in pre-training. The best and the second best results are boldfaced and underlined, respectively.}
\label{tab:result:size}
\begin{tabular}{l|c|ccc|ccc}
\hline
\multirow{2}{*}{Methods} & {Synthetic} & \multicolumn{3}{c|}{PanoContext} & \multicolumn{3}{c}{2D-3D-S}\\ \cline{3-8}
& {Data Size} & {3D IoU (\%) $\uparrow$} & {CE (\%) $\downarrow$} & {PE (\%) $\downarrow$} & {3D IoU (\%) $\uparrow$} & {CE (\%) $\downarrow$} & {PE (\%) $\downarrow$} \\ 
\hline
\multirow{3}{*}{LayoutNet~\cite{ZouCSH18,ZouSPCSWCH19}}
& 1k & 83.81 & \underline{0.66} & 1.99 & 83.57 & 0.72 & 2.31\\ 
& 5k & \underline{84.47} & {0.67} & \underline{1.97} & {\bf 84.55} & \underline{0.69} & \underline{2.21} \\
& 10k & {\bf 84.77} & {\bf 0.63} & {\bf 1.89} & \underline{84.04} & {\bf 0.66} & {\bf 2.08} \\
\hline
\multirow{3}{*}{HorizonNet~\cite{SunHSC19}}
& 1k & 83.77 & 0.74 & 2.11 & 85.19 & \underline{0.63} & 2.01 \\
& 5k & \underline{84.13} & \underline{0.73} & \underline{2.07} & {\bf 86.35} & {\bf 0.61} & \underline{1.87} \\
& 10k & {\bf 85.27} & {\bf 0.66} & {\bf 1.86} & \underline{86.01} & {\bf 0.61} & {\bf 1.84} \\
\hline
\end{tabular}
\end{table}

\smallskip
\noindent{\bf Performance vs. synthetic data size.} We further study the relationship between the number of synthetic images used in pre-training and the accuracy on the real dataset. We sample 1k, 5k and 10k synthetic images for pre-training, then fine-tune the model on the real dataset. The results are shown in \tabref{tab:result:size}. As expected, using more synthetic data generally improves the performance.

\begin{table}[t]
\scriptsize
\renewcommand{\arraystretch}{1.2}
\centering
\caption{Domain adaptation results. NA: non-adaptive baseline. +DA: align layout estimation output. +Depth: align both layout estimation and depth outputs. Real: train in the target domain.}
\label{tab:result:adaptation}
\begin{tabular}{l|ccc|ccc}
\hline
\multirow{2}{*}{Methods} & \multicolumn{3}{c|}{PanoContext} & \multicolumn{3}{c}{2D-3D-S} \\ \cline{2-7}
& {3D IoU (\%) $\uparrow$} & {CE (\%) $\downarrow$} & {PE (\%) $\downarrow$} & {3D IoU (\%) $\uparrow$} & {CE (\%) $\downarrow$} & {PE (\%) $\downarrow$} \\
\hline
NA & 75.64 & 1.31 & 4.10 & 57.18 & 2.28 & 7.55 \\
+DA & 76.91 & 1.19 & 3.64 & 70.08 & 1.36 & 4.66 \\
+Depth & 78.34 & 1.03 & 2.99 & 72.99 & 1.24 & 3.60 \\
Real & 81.76 & 0.95 & 2.58 & 81.82 & 0.96 & 3.13 \\
\hline
\end{tabular}
\end{table}

\smallskip
\noindent{\bf Domain adaptation.} Domain adaptation techniques (\eg, \cite{TsaiHSSYC18}) have been shown to be effective in bridging the performance gap when directly applying models learned on synthetic data to real environments. In this experiment, we do not assume access to ground truth layout labels in the real dataset. We adopt LayoutNet as the task network and use PanoContext and 2D-3D-S separately. We apply a discriminator network to align the output features of the LayoutNet for two domains. Inspired by~\cite{ChenLCVG19}, we further leverage multi-modal annotations in our dataset by adding another decoder branch to the LayoutNet for depth prediction. We concatenate the boundary, corner, and depth predictions as the input of the discriminator network. The results are shown in the \tabref{tab:result:adaptation}. By incorporating additional information, \ie, depth map, we further boost the performance on both datasets. This illustrates the advantage of including multiple types of ground truth in our dataset.

\begin{figure}[t]
\centering
\begin{tabular}{cc}
\includegraphics[width=0.45\linewidth]{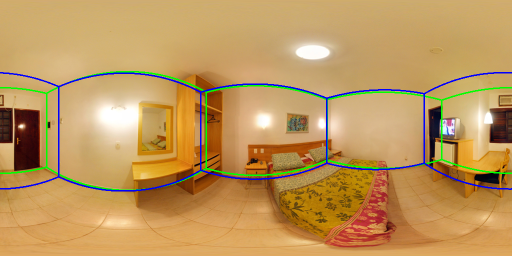} & 
\includegraphics[width=0.45\linewidth]{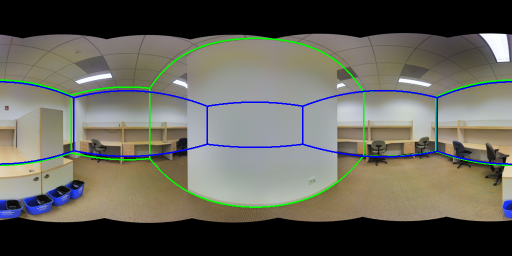}
\end{tabular}
\caption{Limitation of real datasets. {\bf Left}: PanoContext dataset. {\bf Right}: 2D-3D-S dataset. Blue lines are ground truth layout and green lines are predictions.}
\label{fig:failure}
\end{figure}

\smallskip
\noindent{\bf Limitation of real datasets.} Due to human errors, the annotation in real datasets is not always consistent with the actual room layout. In the left image of \figref{fig:failure}, the room is a non-cuboid layout, but the ground truth layout is labeled as cuboid shape. In the right image, the front wall is not labeled as ground truth. These examples illustrate the limitation of using real datasets as benchmarks. We avoid such errors in our dataset by automatically generating ground truth from the original design files.

\section{Conclusion}

In this paper, we present Structured3D, a large synthetic dataset with rich ground truth 3D structure annotations of 21,835 rooms and more than 196k photo-realistic 2D renderings. Among many potential use cases of our dataset, we further demonstrate its benefit in augmenting real data and facilitating domain adaptation for the room layout estimation task.

We view this work as an important and exciting step towards building intelligent machines which can achieve human-level holistic 3D scene understanding. 
In the future, we will continue to add more 3D structure annotations of the scenes and objects to the dataset, and explore novel ways to use the dataset to advance techniques for structured 3D modeling and understanding.

\smallskip
\noindent{\bf Acknowledgement.} We would like to thank Kujiale.com for providing the database of house designs and the rendering engine. We especially thank Qing Ye and Qi Wu from Kujiale.com for the help on the data rendering. This work was partially supported by the National Key R\&D Program of China (\#2018AAA0100704) and the National Science Foundation of China (\#61932020). Zihan Zhou was supported by NSF award \#1815491.

\bibliographystyle{splncs04}
\bibliography{references}
\end{document}